\documentclass[sigconf]{acmart}

\usepackage{balance}

\def\BibTeX{{\rm B\kern-.05em{\sc i\kern-.025em b}\kern-.08emT\kern-.1667em\lower.7ex\hbox{E}\kern-.125emX}}

\copyrightyear{2021} 
\acmYear{2021} 
\setcopyright{acmcopyright}
\acmConference[SIGIR '21]{Proceedings of the 44th International ACM SIGIR Conference on Research and Development in Information Retrieval}{July 11--15, 2021}{Virtual Event, Canada}
\acmBooktitle{Proceedings of the 44th International ACM SIGIR Conference on Research and Development in Information Retrieval (SIGIR '21), July 11--15, 2021, Virtual Event, Canada}
\acmPrice{15.00}
\acmDOI{10.1145/3404835.3462858}
\acmISBN{978-1-4503-8037-9/21/07}

\usepackage{multirow}
\usepackage{bbding}
\usepackage{subfigure}
\begin{document}

\fancyhead{}

\title{Partner Matters! An Empirical Study on Fusing Personas for Personalized Response Selection in Retrieval-Based Chatbots}


\author{Jia-Chen Gu$^1$, Hui Liu$^2$, Zhen-Hua Ling$^1$, Quan Liu$^{1,3}$, Zhigang Chen$^3$, Xiaodan Zhu$^2$}
\affiliation{\institution{$^1$National Engineering Laboratory for Speech and Language Information Processing, \\ University of Science and Technology of China, Hefei, China}}
\affiliation{\institution{$^2$ECE \& Ingenuity Labs, Queen's University, Kingston, Canada}}
\affiliation{\institution{$^3$State Key Laboratory of Cognitive Intelligence, iFLYTEK Research, Hefei, China}}
\email{gujc@mail.ustc.edu.cn, {hui.liu, xiaodan.zhu}@queensu.ca, {zhling, quanliu}@ustc.edu.cn, zgchen@iflytek.com}








\begin{abstract}
  \emph{Persona} can function as the prior knowledge for maintaining the consistency of dialogue systems.
  Most of previous studies adopted the \emph{self} persona in dialogue whose response was about to be selected from a set of candidates or directly generated, but few have noticed the role of \emph{partner} in dialogue.
  This paper makes an attempt to thoroughly explore the impact of utilizing personas that describe either \emph{self} or \emph{partner} speakers on the task of response selection in retrieval-based chatbots.
  Four persona fusion strategies are designed, which assume personas interact with \emph{contexts} or \emph{responses} in different ways.
  These strategies are implemented into three representative models for response selection, which are based on the Hierarchical Recurrent Encoder (HRE), Interactive Matching Network (IMN) and Bidirectional Encoder Representations from Transformers (BERT) respectively.
  Empirical studies on the \texttt{Persona-Chat} dataset show that the partner personas neglected in previous studies can improve the accuracy of response selection in the IMN- and BERT-based models.
  Besides, our BERT-based model implemented with the context-response-aware persona fusion strategy outperforms previous methods by margins larger than 2.7\% on original personas and 4.6\% on revised personas in terms of $\textbf{hits}@1$ (top-1 accuracy), achieving a new state-of-the-art performance on the \texttt{Persona-Chat} dataset.
\end{abstract}

%
%

\ccsdesc[500]{Information systems}
\ccsdesc[500]{Information systems~Personalization}

%
\keywords{Persona fusion, self and partner speakers, multi-turn response selection, retrieval-based chatbot}

%

%
\maketitle

\section{Introduction}
  Building human-like conversational systems has been a long-standing goal in artificial intelligence, where one of the major challenges is to present a consistent personality \cite{zheng2019personalized}.
  Personalized response selection, which aims to select an appropriate response from a set of candidates given the conversation context and personas of speakers, is an important technique to present personalities of dialogue agents in retrieval-based chatbots \cite{zhang2018personalizing,mazare2018training,zhao2019document,gu2019dually,gu2020filtering}. 
  Retrieval-based chatbots are commonly used to build dialogue agents \cite{lowe2015ubuntu,wu2017sequential,zhang2018modeling,zhou2018multi,tao2019multi,tao2019one,gu2019interactive,gu2020utterance,yuan2019multi,gu2020speaker,gu2021deep}. 
  Nowadays, many companies have built retrieval-based virtual assistants due to its promising potentials and alluring commercial values \cite{kepuska2018next,berdasco2019user,zhou2020design}. 
  \citet{zhang2018personalizing} constructed a \texttt{Persona-Chat} dataset for building personalized dialogue agents, where each persona was represented as multiple sentences of profile description.
  A raw example dialogue conditioned on given profiles from the \texttt{Persona-Chat} dataset is shown in Table~\ref{tab4}.

  \begin{table*}[t]
     \caption{An example dialogue from the \texttt{Persona-Chat} dataset.}
     \centering
     \begin{tabular}{c|l|c|l}
      \toprule
       \multicolumn{2}{c|}{\textbf{Persona 1}} & \multicolumn{2}{c}{\textbf{Persona 2}} \\
      \hline
       \multirow{5}{*}{Original} & I just bought a brand new house.   & \multirow{5}{*}{Original} &  I love to meet new people. \\
                                 & I like to dance at the club.       &                           &  I have a turtle named timothy. \\
                                 & I run a dog obedience school.      &                           &  My favorite sport is ultimate frisbee. \\
                                 & I have a big sweet tooth.          &                           &  My parents are living in bora bora. \\
                                 & I like taking and posting selkies. &                           &  Autumn is my favorite season. \\
      \hline
       \multirow{5}{*}{Revised}  & I have purchased a home.                     & \multirow{5}{*}{Revised} & I like getting friends. \\
                                 & Just go dancing at the nightclub, it is fun! &                          & Reptiles make good pets. \\
                                 & I really enjoy animals.                      &                          & I love to run around and get out my energy. \\
                                 & I enjoy chocolate.                           &                          & My family lives on a island. \\
                                 & I pose for pictures and put them online.     &                          & I love watching the leaves change colors. \\
      \hline
       \multicolumn{4}{l}{\textbf{Dialogue}} \\
       \multicolumn{4}{l}{Person 1: Hello, how are you doing tonight?} \\
       \multicolumn{4}{l}{Person 2: I am well an loving this interaction how are you?} \\
       \multicolumn{4}{l}{Person 1: I am great. I just got back from the club.} \\
       \multicolumn{4}{l}{Person 2: This is my favorite time of the year season wise.} \\
       \multicolumn{4}{l}{Person 1: I would rather eat chocolate cake during this season.} \\
       \multicolumn{4}{l}{Person 2: What club did you go to? Me an timothy watched tv.} \\
       \multicolumn{4}{l}{Person 1: I went to club chino. What show are you watching?} \\
       \multicolumn{4}{l}{Person 2: LOL oh okay kind of random.} \\
       \multicolumn{4}{l}{Person 1: I love those shows. I am really craving cake.} \\
       \multicolumn{4}{l}{Person 2: Why does that matter any? I went outdoors to play frisbee.} \\
       \multicolumn{4}{l}{Person 1: It matters because I have a sweet tooth.} \\
       \multicolumn{4}{l}{Person 2: So? LOL I want to meet my family at home in bora.} \\
       \multicolumn{4}{l}{Person 1: My family lives in alaska. It is freezing down there.}\\
       \multicolumn{4}{l}{Person 2: I bet it is oh I could not.} \\
      \bottomrule
      \end{tabular}
      \label{tab4}
    \end{table*}

  Although great progress has been made for building personalized dialogue agents \cite{li2016persona,zhang2018personalizing,mazare2018training,zhao2019document,gu2019dually,gu2020filtering}, they are still in their infancy.
  Most of previous studies focused on the self speaker's persona in dialogue who was about to utter a response, while the contribution of the partner speaker's persona to dialogue was rarely noticed. 
  For a conversation conditioned on personas, if a dialogue agent has no access to the partner persona, it often over-focuses on retrieving responses related to the agent itself, which sometimes deviates from the ground truth of how a conversation really goes. 
  For example, given a conversation about hobbies, if the agent only has access to the self persona profile ``I like playing basketball'', it often over-weights response candidates such as “I like sports”. 
  However, if the agent also has access to the partner persona profile ``I often play various instruments'', it gives models more flexibility to not only focus on continuously talking about the agent itself, but also conducting more collaborative communication, e.g., asking questions such as ``Who is your favorite musician'', as the real conversations often proceed. 
  
  Whether and how personas of different speakers in dialogue contribute to building coherent personalized dialogue models is a fundamental problem.
  In order to compare the ability of different personas for selecting an appropriate response directly, we first perform the preliminary experiments by ablating the context information and searching for an appropriate response with only the given self or partner persona information.
  Hierarchical Recurrent Encoder (HRE) \cite{serban2016building}, Interactive Matching Network (IMN) \cite{gu2019interactive} and Bidirectional Encoder Representations from Transformers (BERT) \cite{devlin2019bert} are chosen as the matching models and the results are shown in Table~\ref{tab6}.
  The results show that the single persona-response matching can achieve a comparable performance, which shows the usefulness of utilizing persona information to select an appropriate response.
  Meanwhile, it can be seen that although the partner persona is less important than the self persona, it can still contribute to response selection to some extent, which is consistent with our assumption mentioned above.
  However, few studies noticed the role of \emph{partner} in dialogue and under what conditions the partner speaker's persona can contribute more has not been studied too much yet.

  \begin{table}[t]
    \caption{Performance of persona-response matching on the \texttt{Persona-Chat} dataset conditioned on the original persona.}
    \centering
    \begin{tabular}{c|c|c|c}
    \toprule
             Model         & Persona & $\textbf{hits}@1$ & \textbf{MRR} \\
    \hline
    \multirow{2}{*}{HRE}   & Self    & 23.9              & 40.1  \\
                           & Partner &  8.7              & 23.4  \\
    \hline
    \multirow{2}{*}{IMN}   & Self    & 48.8              & 60.7  \\
                           & Partner & 19.3              & 34.2  \\
    \hline
    \multirow{2}{*}{BERT}  & Self    & 50.6              & 62.5  \\
                           & Partner & 20.6              & 35.6  \\

    \bottomrule
    \end{tabular}
    \label{tab6}
  \end{table}

  To this end, we make an attempt to explore the impact of utilizing personas that describe either \emph{self} or \emph{partner} speakers on the task of personalized response selection.
  Four persona fusion strategies, i.e., none-aware (NA), context-aware (CA), response-aware (RA) and context-response-aware (CRA) ones, are designed based on whether or not considering the interactions between personas and contexts as well as the interactions between personas and responses.
  For a thorough comparison and analysis, these four strategies are implemented into three representative models for response selection, which are based on the HRE, IMN and BERT models respectively.
  HRE follows the sentence-encoding-based framework for response selection, which encodes contexts and responses separately without interactions between them and obtains their embeddings separately.
  As a representative model under the cross-attention-based framework, IMN performs the interactive matching between contexts and responses to derive the matching information between them.
  Meanwhile, IMN shares the most similar architecture with HRE we implemented in this paper, so that we can explore the effect of interactions between contexts and responses on persona fusion.
  The BERT-based response selection model not only performs interactions between contexts and responses, but also incorporates generic distributional semantics and other knowledge through pretraining.

  We introduce our models and test the proposed persona fusion methods on the \texttt{Persona-Chat} dataset \cite{zhang2018personalizing} which is the largest public dataset to date containing multi-turn dialogues conditioned on personas.
  Experimental results show that the partner persona contributes to the performance when using the IMN- and BERT-based models.
  Besides, the pretraining algorithms can help to capture deep semantics given more context.
  Furthermore, compared with previous methods, our BERT-based model implemented with the context-response-aware persona fusion strategy improves $\textbf{hits}@1$ (top-1 accuracy) by 2.7\% on original personas and by 4.6\% on revised personas, achieving a new state-of-the-art performance on this dataset. 

  In summary, the contributions of this paper are two-fold.
  First, four persona fusion strategies are designed and implemented into three models, aiming to explore the impact of utilizing the personas of not only \emph{self} but also \emph{partner} speakers on response selection.
  Second, experimental results demonstrate that our proposed models outperform the existing state-of-the-art models by large margins on the widely used \texttt{Persona-Chat} response selection benchmark.

\section{Related Work}

  Chit-chat models suffer from a lack of a consistent personality as they are typically trained over many dialogues, each with different speakers, and a lack of explicit long-term memory as they are typically trained to produce an utterance given only a very recent dialogue history.
  Existing methods used to build a dialogue agents can be generally categorized into generation-based \cite{li2016persona,serban2016building,serban2017hierarchical} and retrieval-based methods \cite{lowe2015ubuntu,wu2017sequential,zhang2018modeling,zhou2018multi,tao2019multi,tao2019one,gu2019interactive,gu2020utterance,yuan2019multi,gu2020speaker,gu2021deep}.
  Nowadays, many companies have built personalized virtual assistants due to its promising potentials and alluring commercial values \cite{kepuska2018next,berdasco2019user,zhou2020design}. 
  \citet{li2016persona} proposed a persona-based neural conversation model to capture individual characteristics such as background information and speaking style.
  \citet{miller2016key} proposed the key-value memory network, where the keys were dialogue histories, i.e., contexts, and the values were next dialogue utterances.
  \citet{zhang2018personalizing} constructed a \texttt{Persona-Chat} dataset for building personalized dialogue agents, which is the largest public dataset to date containing multi-turn dialogues conditioned on personas.
  It also established many baselines for this benchmark, such as the profile memory network by considering the dialogue history as input and then performing attention over the persona to be combined with the dialogue history.
  \citet{mazare2018training} proposed the fine-tuned \texttt{Persona-Chat} (FT-PC) model which first pretrained models using a large-scale corpus based on Reddit to extract valuable dialogues conditioned on personas, and then fine-tuned these pretrained models on the \texttt{Persona-Chat} dataset.
  \citet{zheng2019personalized} proposed to incorporate explicit personality traits, such as age, gender and location, into conversation.
  \citet{luo2019learning} proposed to combine a profile model and a preference model into the personalized MEMN2N which encodes user profiles into distributed embeddings and refers to conversation history from other similar users. Then a PREFERENCE MODEL captures user preferences over knowledge base entities to handle the ambiguity in user requests. 
  \citet{zhao2019document} proposed a document-grounded matching network (DGMN) which fused information in a persona and a context into representations of each other, and dynamically determined if persona information is necessary and the importance of different parts of a persona and a context.
  \citet{gu2019dually} proposed a dually interactive matching network (DIM) for presenting the personalities of dialogue agents by performing the interactive matching between responses and contexts and between responses and personas respectively for ranking response candidates.
  \citet{wolf2019transfertransfo} and \citet{liu2020you} both employed the pre-trained language model of Generative Pretrained Transformer (GPT) \cite{radford2018improving} for building personalized dialogue agents.
  \citet{gu2020filtering} proposed filtering before iteratively referring (FIRE) to ground the conversation on the given knowledge and then perform the deep and iterative matching.

  In general, most of these methods adopted the self persona in dialogue.
  To the best of our knowledge, only \citet{zhang2018personalizing} who constructed the \texttt{Persona-Chat} dataset established several baselines for leveraging the partner persona and \citet{gu2019dually} just tested their method under the partner persona setting.
  No follow-up work explored how to leverage the partner persona to improve the performance under certain conditions.
  Thus, this paper makes an attempt to thoroughly explore the impact of utilizing the personas of not only self but also partner speakers on the performance of personalized response selection. 
  It should be emphasized that the focus of this paper is not so much on designing drastically new models, but instead on investigating the conditions under which the self and partner personas can work. 
  We aim to comprehensively understand the impact of utilizing the personas from not only the self but also the partner speakers on personalized response selection. 
  Thus, we design four persona fusion strategies, and choose three very representative models to apply these strategies into these models in order to verify the effectiveness of these strategies. 
  We can certainly choose other models because models are just a testbed for applying these strategies, which are not the focus of this paper. 
  Instead, exploring the conditions under which the self and partner personas can work is our focus.
  We hope our work help shed some light on combining self and partner personas to further improve response selection performance.

\section{Task Definition}
  Given a dialogue dataset $\mathcal{D}$ with personas, an example of the dataset can be represented as tuple $(c,p,r,y)$ and is shown in Table \ref{tab4}.
  Specifically, $c = \{u_1,u_2,...,u_{n_c}\}$ represents a context with $\{u_m\}_{m=1}^{n_c}$ as its utterances and $n_c$ as the utterance number.
  $p = \{p_1,p_2,...,p_{n_p}\}$ represents a persona with $\{p_n\}_{n=1}^{n_p}$ as its profile sentences and $n_p$ as the profile number.
  $r$ represents a response candidate.
  $y \in \{0,1\}$ denotes a label.
  $y=1$ indicates that $r$ is a proper response for $(c,p)$; otherwise, $y=0$.
  Our goal is to learn a matching model $g(c,p,r)$ from $\mathcal{D}$.
  For any context-persona-response triple $(c,p,r)$, $g(c,p,r)$ measures the matching degree between $(c,p)$ and $r$.

\section{Persona Fusion for Response Selection}

  Capturing personas of different speakers in dialogue is the key for developing personalized dialogue agents.
  In order to thoroughly explore the impact of both self and partner personas on dialogue, we design four persona fusion strategies that assume personas interact with \emph{contexts} or \emph{responses} in different ways and implement them into three models, which are sentence-encoding-based, cross-attention-based and pretraining-based ones.
  Details about the model structures and the strategies are presented in this section.

  \begin{figure}[t]
    \centering
    \subfigure[HRE]{
    \includegraphics[width=4cm]{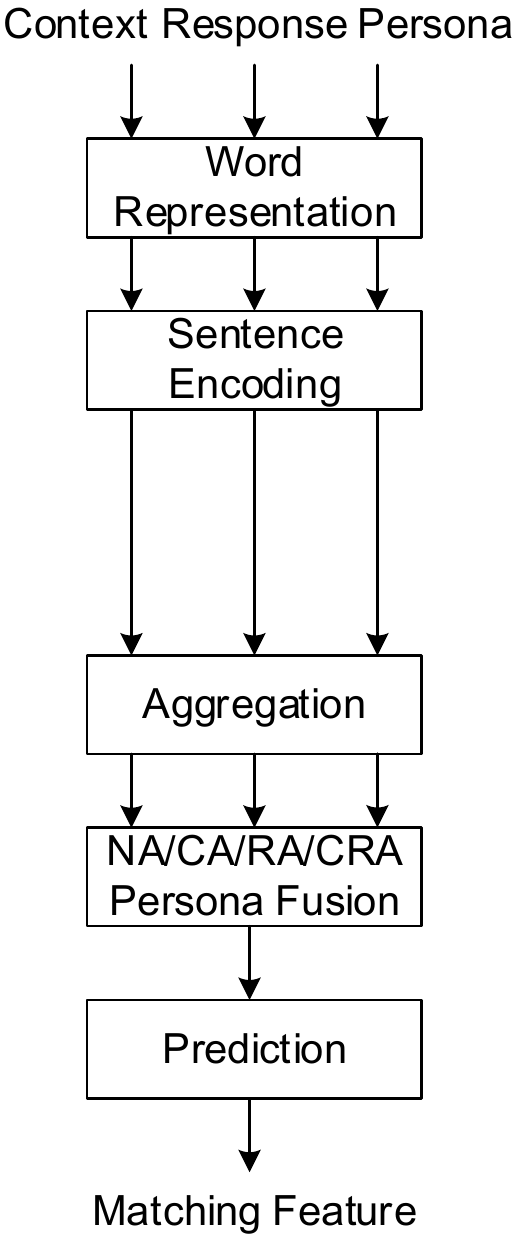}}
    \subfigure[IMN]{
    \includegraphics[width=4cm]{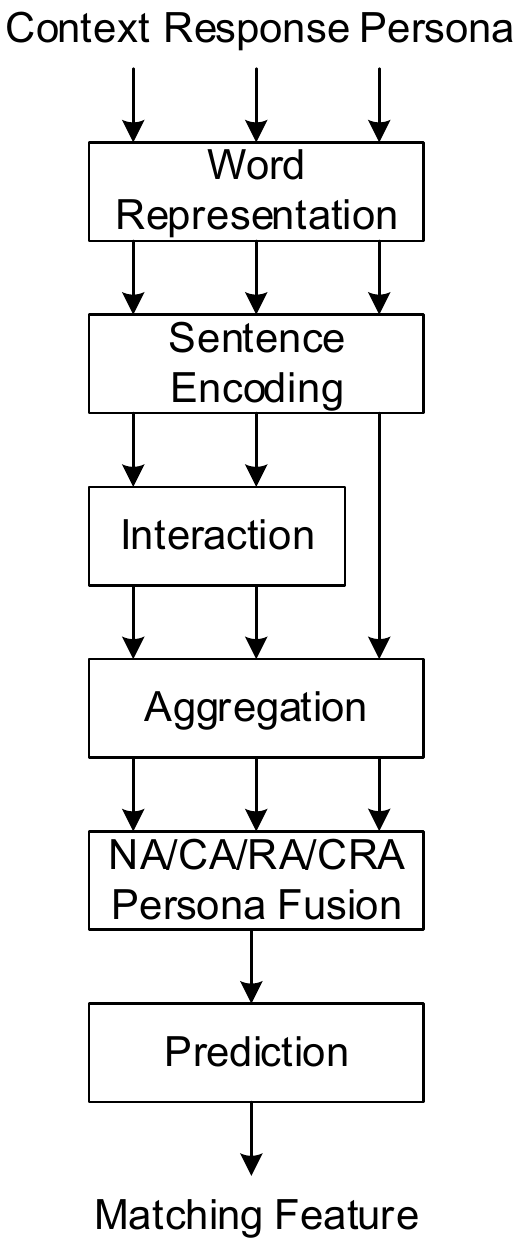}}
    \caption{Comparison of the model architectures for (a) HRE and (b) IMN.}
    \label{fig1}
  \end{figure}

  \subsection{Sentence-Encoding-Based Model}  \label{sec1}
    A representative model under the sentence-encoding-based framework for multi-turn dialogue is Hierarchical Recurrent Encoder-Decoder (HRED) \cite{serban2016building} which was originally proposed for dialogue generation.
    Here, we only need the encoder part to obtain the encoded embedding, so we name it Hierarchical Recurrent Encoder (HRE) in this paper.
    
    Figure~\ref{fig1} (a) shows an overview of the architecture.
    First, we follow the setting used in IMN \cite{gu2019interactive}, which constructs word representations by combining general pretrained word embeddings, those estimated on the task-specific training set, as well as character-level embeddings, in order to deal with the out-of-vocabulary issue.
    Formally, embeddings of the \emph{m}-th utterance in a context, the \emph{n}-th profile sentence in a persona and a response candidate are denoted as $\textbf{U}_m = \{\textbf{u}_{m,i}\}_{i=1}^{l_{u_m}}$, $\textbf{P}_n = \{\textbf{p}_{n,j}\}_{j=1}^{l_{p_n}}$ and $\textbf{R} = \{\textbf{r}_k\}_{k=1}^{l_r}$ respectively, where $l_{u_m}$, $l_{p_n}$ and $l_r$ are the numbers of words in $\textbf{U}_m$, $\textbf{P}_n$ and $\textbf{R}$ respectively.
    Each $\textbf{u}_{m,i}$, $\textbf{p}_{n,j}$ or $ \textbf{r}_k$ is an embedding vector. 

    Then, context utterances, persona profiles and response candidate are encoded by bidirectional long short-term memories (BiLSTMs) \cite{hochreiter1997long}.
    Detailed calculations of BiLSTM are omitted due to limited space.
    We denote the calculations as
    \begin{align}
      \bar{\textbf{u}}_{m,i} &= \textbf{BiLSTM}(\textbf{U}_m, i), i \in \{1, ..., l_{u_m}\},\\
      \bar{\textbf{p}}_{n,j} &= \textbf{BiLSTM}(\textbf{P}_n, j), j \in \{1, ..., l_{p_n}\},\\
      \bar{\textbf{r}}_{k}   &= \textbf{BiLSTM}(\textbf{R}, k),   k \in \{1, ..., l_{r}\},
    \end{align}
    where $\bar{\textbf{U}}_m = \{\bar{\textbf{u}}_{m,i}\}_{i=1}^{l_{u_m}}$, $\bar{\textbf{P}}_n = \{\bar{\textbf{p}}_{n,j}\}_{j=1}^{l_{p_n}}$ and $\bar{\textbf{R}} = \{\bar{\textbf{r}}_k\}_{j=1}^{l_r}$.
    The parameters of these three BiLSTMs are shared in our implementation.
    Each $\bar{\textbf{u}}_{m,i}$, $\bar{\textbf{p}}_{n,j}$, or $\bar{\textbf{r}}_k$ is an embedding vector.

    The matching matrix $\bar{\textbf{U}}_m$, $\bar{\textbf{P}}_n$, and $\bar{\textbf{R}}$ are aggregated by the max and last-hidden-state pooling operations to derive their embedding vectors as
    \begin{align}
      \bar{\textbf{u}}_m^{agr} &= [\bar{\textbf{u}}_{m,max};\bar{\textbf{u}}_{m,l_{u_m}}], m \in \{1, ..., n_c\},\\
      \bar{\textbf{p}}_n^{agr} &= [\bar{\textbf{p}}_{n,max};\bar{\textbf{p}}_{n,l_{p_n}}], n \in \{1, ..., n_p\},\\
      \bar{\textbf{r}}^{agr}   &= [\bar{\textbf{r}}_{max};  \bar{\textbf{r}}_{l_r}].
    \end{align}

    Next, the sequences of $\bar{\textbf{u}}_m^{agr}$ and $\bar{\textbf{p}}_n^{agr}$ are further aggregated to get the embedding vectors for the context and the persona respectively.
    As the utterances in a context are chronologically ordered, the utterance embeddings $\bar{\textbf{U}}^{agr}=\{\bar{\textbf{u}}_{m}^{agr}\}_{m=1}^{n_c}$ are sent into another BiLSTM following the chronological order of utterances in the context.
    Combined max pooling and last-hidden-state pooling operations are then performed to obtain the context embeddings as
    \begin{align}
      \hat{\textbf{u}}_m = \textbf{BiLSTM}(\bar{\textbf{U}}^{agr}, m)&, m \in \{1, ..., n_c\},\\
      \hat{\textbf{c}}^{agr} = [\hat{\textbf{u}}_{max};\hat{\textbf{u}}_{n_c}]&.
    \end{align}

    Similarly, given the sequence of profile embeddings $\{\bar{\textbf{p}}_n^{agr}\}_{n=1}^{n_p}$, the aggregated persona embedding $\hat{\textbf{p}}^{agr}$ is obtained by persona fusion.
    In this paper, four persona fusion strategies are designed based on whether or not considering the interactions between personas and contexts, and the interactions between personas and responses.

    \subsubsection{None-Aware Persona Fusion}
      In this strategy, the persona fusion is independent of both contexts and responses.
      A self-attention-based aggregation is designed to derive the persona embedding as follows,
      \begin{align}
        \alpha_n               &= \textbf{w}^\top \cdot \bar{\textbf{p}}_n^{agr} + b,\\
        \hat{\textbf{p}}^{agr} &= \sum_{n=1}^{n_p} \frac{e^{\alpha_n}} {\sum_{k=1}^{n_p} e^{\alpha_k}} \bar{\textbf{p}}_n^{agr},
        \label{equ2}
      \end{align}
      where $\textbf{w}$ and $b$ are parameters that need to be estimated during training.
      Then, the aggregated persona embedding is fused as a part of the final matching feature as shown in Eq.~(\ref{equ1}).
      This persona fusion strategy is not aware of any information of context and response, so we name it accordingly as \emph{none-aware} (NA) persona fusion in this paper.

    \subsubsection{Context-Aware Persona Fusion}
      In order to make aware of the context information when fusing the persona, we design a context-aware (CA) persona fusion strategy by computing similarities between the context embedding and each profile embedding, and then performing the attention operation to obtain the aggregated persona embedding $\hat{\textbf{p}}^{agr}$ as
      \begin{align}
        \alpha_n               &= \hat{\textbf{c}}^{{agr}^\top} \cdot \bar{\textbf{p}}_n^{agr}, \\
        \hat{\textbf{p}}^{agr} &= \sum_{n=1}^{n_p} \frac{e^{\alpha_n}} {\sum_{k=1}^{n_p} e^{\alpha_k}} \bar{\textbf{p}}_n^{agr},
      \end{align}
      This persona fusion strategy is aware of the context information by attaching different importance to profile embeddings according to their similarities to the context dynamically, so we name it \emph{context-aware} (CA) persona fusion in this paper.

    \subsubsection{Response-Aware Persona Fusion}
      Similarly, we design a \emph{response-aware} (RA) persona fusion strategy by computing similarities between the response embedding and each profile embedding, and then performing the attention operation to obtain the aggregated persona embedding $\hat{\textbf{p}}^{agr}$ as follows,
      \begin{align}
        \alpha_n               &= \bar{\textbf{r}}^{{agr}^\top} \cdot \bar{\textbf{p}}_n^{agr}, \\
        \hat{\textbf{p}}^{agr} &= \sum_{n=1}^{n_p} \frac{e^{\alpha_n}} {\sum_{k=1}^{n_p} e^{\alpha_k}} \bar{\textbf{p}}_n^{agr},
      \end{align}
      Then the same attention operation as Eq.~(\ref{equ2}) is performed to obtain $\hat{\textbf{p}}^{agr}$.

    \subsubsection{Context-Response-Aware Persona Fusion}
      In order to make aware of both the context and the response information simultaneously, we design a \emph{context-response-aware} (CRA) persona fusion strategy.
      This strategy first concatenates the context and the response embedding, and then transforms it to the same dimension of profile embeddings with a linear transformation. 
      Similarities are computed between it and each profile embedding. 
      Then the same attention operation is performed to obtain $\hat{\textbf{p}}^{agr}$.
      Mathematically, we have
      \begin{align}
        \alpha_n               &= ( \textbf{w}^\top \cdot [\hat{\textbf{c}}^{agr};\bar{\textbf{r}}^{agr}] +b )^\top \cdot \bar{\textbf{p}}_n^{agr}, \\
        \hat{\textbf{p}}^{agr} &= \sum_{n=1}^{n_p} \frac{e^{\alpha_n}} {\sum_{k=1}^{n_p} e^{\alpha_k}} \bar{\textbf{p}}_n^{agr}.
      \end{align}

    Lastly, after obtaining the aggregated persona embedding $\hat{\textbf{p}}^{agr}$, the final matching feature vector is the concatenation of the context, persona and response embeddings as
    \begin{align}
      \textbf{m} = [\hat{\textbf{c}}^{agr};\hat{\textbf{p}}^{agr};\bar{\textbf{r}}^{agr}].
      \label{equ1}
    \end{align}
    The final matching feature vector is then sent into a multi-layer perceptron (MLP) classifier.
    Here, the MLP classifier is designed to predict whether a context-response-persona triple $(c,p,r)$ match appropriately based on the derived matching feature vector and returns a score denoting the matching degree of this triple.
    Finally, a softmax output layer is adopted in the MLP to return a probability distribution over all response candidates.
    Models are learnt by minimizing the MLP cross-entropy loss.
    Let $\Theta$ denote the model parameters.
    The learning objective $\mathcal{L}(\mathcal{D}, \Theta)$ is formulated as
    \begin{equation}
      \mathcal{L}(\mathcal{D}, \Theta) = - \sum_{(c,p,r,y)\in \mathcal{D}} y log(g(c,p,r)) .
    \end{equation}

  \subsection{Cross-Attention-Based Model}
    A representative model under the cross-attention-based framework for multi-turn dialogue is Interactive Matching Network (IMN) \cite{gu2019interactive}, which was originally proposed for multi-turn response selection.
    Another reason to choose this model is that it shares the most similar architecture with HRE we implemented in this paper, so that we can explore the effect of interactions between contexts and responses on persona fusion.
    
    Figure~\ref{fig1} (b) shows an overview of the architecture. 
    IMN shares with HRE the same modules of word representation, sentence encoding, aggregation, persona fusion and prediction.
    In addition, IMN is equipped with an interaction module which performs the global and bidirectional cross-attention operation between the context and the response to capture the matching information between them.
    We introduce the interaction module briefly as follows and readers could refer to \cite{gu2019interactive} for more details of IMN. 
    First, after looking up the wording embedding table and encoded by the sentence encoder to derive the set of utterance representations $\{\bar{\textbf{U}}_m\}_{m=1}^{n_c}$ and the response representations $\bar{\textbf{R}}$, the context representation $\bar{\textbf{C}} = \{\bar{\textbf{c}}_i\}_{i=1}^{l_c} $ with $l_c = \sum_{m=1}^{n_c} l_{u_m}$ is formed by concatenating the set of utterance representations $\{\bar{\textbf{U}}_m\}_{m=1}^{n_c}$. 
    Then, IMN matches the response with the whole context in a global and bidirectional way, i.e., considering the whole context as a single sequence.
    The global context-response matching can help select the most relevant parts of the whole context and neglect the irrelevant parts.
    An attention-based alignment is employed to collect information between the context and the response by computing the attention weight between each $(\bar{\textbf{c}}_i, \bar{\textbf{r}}_k)$ tuple as
    \begin{align}
      e_{ik} = (\bar{\textbf{c}}_i)^T \cdot \bar{\textbf{r}}_k.
    \end{align}
    For a word in the response, its response-to-context relevant representation is composed as a weighted summation of $\{\bar{\textbf{c}}_i\}_{i=1}^{l_c}$. 
    The same calculation is performed for each word in a context to compose its context-to-response representation as a weighted summation of $\{\bar{\textbf{r}}_k\}_{k=1}^{l_r}$. 
    To further enhance the collected information, the element-wise differences and products with their representations after the sentence encoder are calculated and are then concatenated to obtain the enhanced representations.
    Finally, the concatenated context representations need to be converted back to separate utterance representations which are sent for further aggregation.


    It is notable that the persona aggregation in IMN is identical to that in HRE.
    Readers can refer to Section~\ref{sec1} for details.

  \subsection{Pretraining-Based Model}
    
    
    \begin{figure}[t]
      \centering
      \includegraphics[width=8.5cm]{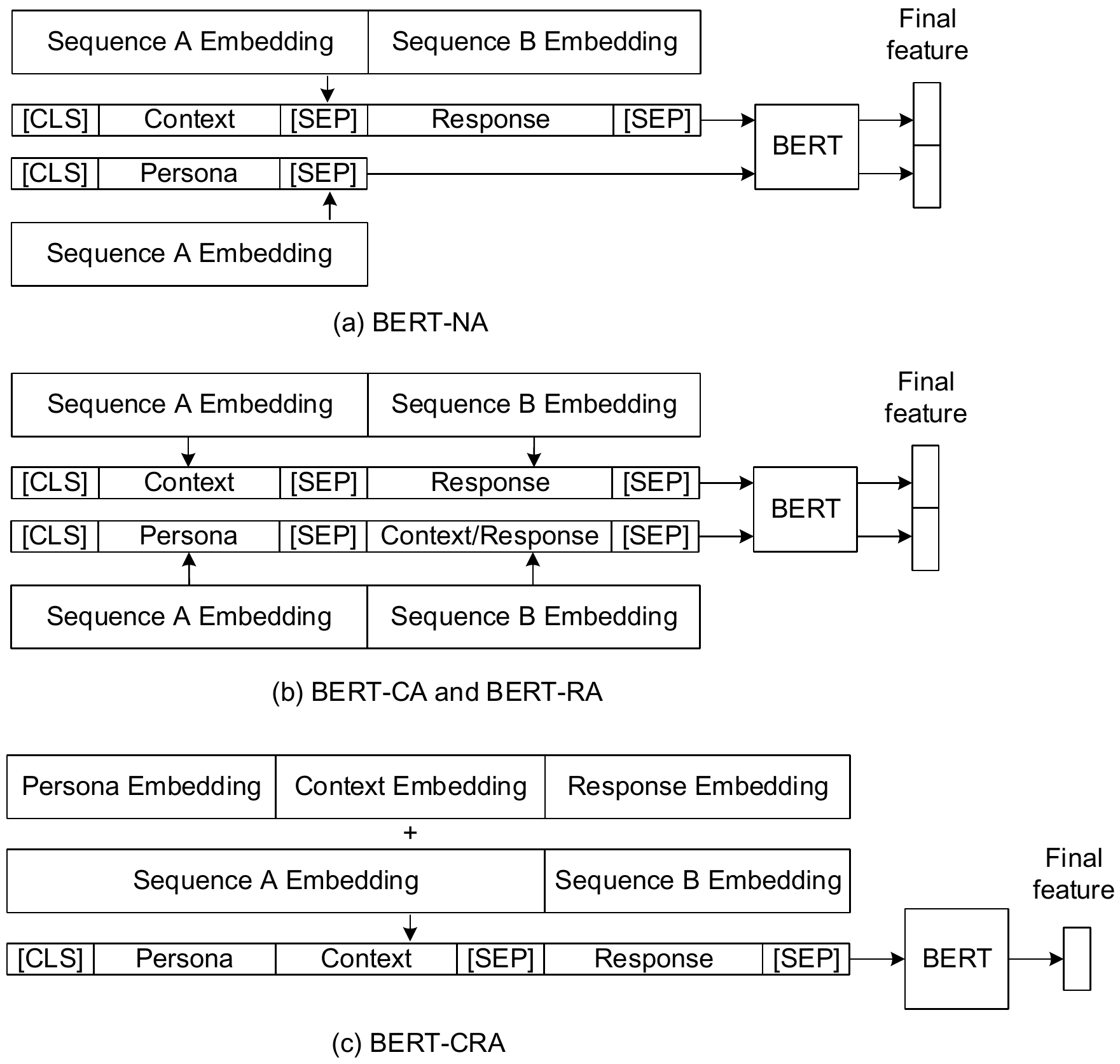}
      \caption{Comparison of the model architectures for (a) BERT-NA, (b) BERT-CA and BERT-RA and (c) BERT-CRA.}
      \label{fig2}
    \end{figure}

    A representative model under the pretraining-based framework is Bidirectional Encoder Representations from Transformers (BERT) \cite{devlin2019bert}.
    Due to space limitation, we omit an exhaustive background description of BERT.
    Readers can refer to \cite{devlin2019bert} for details.
    The four persona fusion strategies are implemented by adjusting BERT to fit the task of personalized response selection in different ways.

    \subsubsection{None-Aware Persona Fusion}
      In this strategy, we propose a dual matching architecture which is composed of two encoding pipelines.
      One is used to derive the matching feature between contexts and responses, and the other is used to derive the persona fusion feature.
      Finally, these two features are concatenated to form the final feature.
      Figure~\ref{fig2} (a) shows an overview of the architecture.

      In order to derive the matching feature between contexts and responses, we follow the configuration in the original BERT that the context is used to form the sequence A in BERT, and the response is used to form the sequence B in BERT.
      Then, these two sequences are concatenated with a \texttt{[SEP]} token to obtain a longer sequence.
      After encoded by stacked Transformer blocks \cite{vaswani2017attention}, the embedding of the first token \texttt{[CLS]} of each concatenated sequence is used as the matching feature for a context-response pair.

      When deriving the persona fusion feature, the persona itself is sent into BERT for encoding without any interactions with contexts or responses. 
      Similarly, the embedding of the first token \texttt{[CLS]} of the persona sequence is used as the persona fusion feature.

      Finally, the matching feature and the persona fusion feature are concatenated to form the final feature.
      This feature captures the matching information in a persona-context-response triple, which is sent into a MLP classifier with a sigmoid output layer.
      The classifier returns a score denoting the matching degree of this triple.

    \subsubsection{Context-Aware Persona Fusion}
      This strategy is similar to the none-aware persona fusion strategy in BERT, except that the persona fusion feature is derived by sending the concatenation of the persona and the context into BERT for encoding.
      Figure~\ref{fig2} (b) shows an overview of the architecture.
      Specifically, the persona is used to form the sequence A in BERT, and the context is used to form the sequence B in BERT.
      Then, these two sequences are concatenated with a \texttt{[SEP]} token.
      Similarly, the embedding of the first token \texttt{[CLS]} is used as the persona fusion feature, which is aware of the context information by interactions between the persona and the context.

    \subsubsection{Response-Aware Persona Fusion}
      The strategy is similar to the context-aware persona fusion strategy in BERT, except that it replaces the context with the response when deriving the persona fusion feature.
      Figure~\ref{fig2} (b) shows an overview of the architecture.

    \subsubsection{Context-Response-Aware Persona Fusion}
      Different from the strategies mentioned above which derive the context-response matching feature and the persona fusion feature respectively, in this strategy, we propose a simple yet effective method to derive a feature which contains both types of information simultaneously.
      Figure~\ref{fig2} (c) shows an overview of the architecture.
      Specifically, the persona and the context are concatenated to form the sequence A, and the response is used to form the sequence B.
      Then these two sequences are concatenated with a \texttt{[SEP]} token.
      In order to further distinguish between them, three subtypes of embedding are added to the corresponding token representations, in addition to the original sequence A/B embeddings, which are parameters updated during the fine-tuning process.
      The encoded embedding of the first token \texttt{[CLS]} of each concatenated sequence is used as the aggregated representation for a persona-context-response triple classification.
      This embedding captures the matching information in this triple.
      In this strategy, the persona fusion is aware of both contexts and responses by interactions with both of them.
      Finally, this emdedding is sent into a MLP classifier, and returns a score denoting the matching degree of this triple.

\section{Experiments}

  \subsection{Dataset}  \label{sec2}
    We tested our proposed methods on the \texttt{Persona-Chat} dataset \cite{zhang2018personalizing} which is the largest public dataset to date containing multi-turn dialogues conditioned on personas.
    The dataset consists of 8939 complete dialogues for training, 1000 for validation, and 968 for testing.
    Response selection is performed at every turn of a complete dialogue, which results in 65719 dialogues for training, 7801 for validation, and 7512 for testing in total.
    Positive responses are true responses from humans and negative ones are randomly sampled by the dataset releaser.
    The ratio between positive and negative responses is 1:19 in the training, validation, and testing sets.
    There are 955 possible personas for training, 100 for validation, and 100 for testing, each consisting of 3 to 5 profile sentences.
    There are no overlaps between the training/validation/testing sets for both dialogues and personas.
    To make this task more challenging, a version of revised persona descriptions are also provided by rephrasing, generalizing, or specializing the original ones.

  \subsection{Evaluation Metrics}
    To ensure results are comparable, we used the same evaluation metrics as in the previous work \cite{zhang2018personalizing,gu2019dually}.
    Each model aimed to select the best-matched response from available candidates for the given context $c$ and persona $p$.
    We calculated the recall of the true positive replies, denoted as $\textbf{hits}@1$.
    In addition, the mean reciprocal rank (\textbf{MRR}) was also adopted to take the rank of the correct response over all candidates into consideration.

  \subsection{Training Details}
    For building HRE, IMN, and their persona fusion models, the ratio of positive and negative responses was set to 1:19 in the training set, with a softmax output layer over all response candidates.
    The Adam method \cite{kingma2014adam} was employed for optimization with a batch size of 16.
    The initial learning rate was 0.001 and was exponentially decayed by 0.96 every 5000 steps.
    Dropout \cite{srivastava2014dropout} with a rate of 0.2 was applied to the word embeddings and all hidden layers.
    The maximum number of training epochs was set to 10.
    The word representation is a concatenation of a 300-dimensional GloVe embedding \cite{pennington2014glove}, a 100-dimensional embedding estimated on the training set using the Word2Vec algorithm \cite{mikolov2013distributed}, and 150-dimensional character-level embeddings with window sizes \{3, 4, 5\}, each consisting of 50 filters.
    The word embeddings were not updated during training.
    All hidden states of the LSTM have 200 dimensions.
    The MLP at the prediction layer have 256 hidden units with ReLU \cite{nair2010rectified} activation.
    The maximum number of characters in a word, that of words in a context utterance, of utterances in a context, of words in a persona profile,  of profiles in a persona, and of words in a response were set to 18, 20, 15, 15, 5, and 20, respectively.
    We padded with zeros if the number of utterances in a context was less than 15; otherwise, we kept the last 15 utterances.
    Similarly, we padded with zeros if the number of profile sentences in a persona was less than 5.
    We used the validation set to select the best model for testing. 

    For building BERT and its persona fusion models, we employed the base version of BERT and most hyper-parameters of the original BERT were followed \cite{devlin2019bert} except the following configurations.
    The initial learning rate was set to 2e-5 and was linearly decayed by L2 weight decay.
    A dynamic negative sampling strategy was adopted that the ratio of positive and negative responses was set to 1:1 in the training set, and it used different negative responses at each epoch.
    Thus, the maximum number of training epochs was set to 19.
    The maximum sequence length was set to 320.
    The training batch size was set to 12.
    The MLP at the prediction layer was a single-layer feed-forward neural network with sigmoid activation.

    All code was implemented in the TensorFlow framework \cite{abadi2016tensorflow} and is published to help replicate our results.\footnote{https://github.com/JasonForJoy/Personalized-Response-Selection}

  \subsection{Comparison Methods}

    \paragraph{Non-pretraining-based methods}
      IR baseline, Starspace, Profile and KV Profile were baselines established in \citet{zhang2018personalizing} who released the \texttt{Persona-Chat} dataset.
      DGMN \cite{zhao2019document}, DIM \cite{gu2019dually} and FIRE \cite{gu2020filtering} were follow-up studies which did not employ any pretraining.

    \paragraph{Pretraining-based methods}
      FT-PC ~\cite{mazare2018training} employed the ``pretrain and fine-tune'' framework by first pretraining on a domain-specific corpus, dialogues of which were extracted from Reddit, and then fine-tuning on the \texttt{Persona-Chat}.
      TransferTransfo \cite{wolf2019transfertransfo} and P$^2$ Bot \cite{liu2020you} were both initialized with the pretrained language model of GPT \cite{radford2018improving} which were pretrained on a large general corpus, and then fine-tuned on the \texttt{Persona-Chat} as well.

  \subsection{Experimental Results}

    \begin{table}[t]
      \caption{Evaluation results of our reimplemented HRE, IMN and BERT models together with previous methods on the \texttt{Persona-Chat} dataset without using any personas.}
      \centering
      \begin{tabular}{l|c|c}
      \toprule
                                                     & $\textbf{hits}@1$ & \textbf{MRR} \\
      \hline
       IR baseline~\cite{zhang2018personalizing}     & 21.4              & -      \\
       Starspace~\cite{zhang2018personalizing}       & 31.8              & -      \\
       Profile~\cite{zhang2018personalizing}         & 31.8              & -      \\
       KV Profile~\cite{zhang2018personalizing}      & 34.9              & -      \\
      \hline
       HRE~\cite{serban2016building}                 & 42.7              & 60.0   \\
       IMN~\cite{gu2019interactive}                  & 63.8              & 75.8   \\
       BERT\cite{devlin2019bert}                     & 70.7              & 80.8   \\
      \bottomrule
      \end{tabular}
      \label{tab1}
    \end{table}

    \begin{table}[t]
     \caption{Performance of four persona fusion strategies implemented into three models on the \texttt{Persona-Chat} dataset under the original persona configuration. 
     Numbers marked with $\star$ denote that the gains or losses after adding persona conditions  are statistically significant (t-test with \emph{p}-value $<$ 0.05) comparing with the corresponding baseline models in Table~\ref{tab1}.
     Numbers in bold denote the persona fusion strategy that achieves the best performance.}
     \centering
     \begin{tabular}{l|c|c|c|c}
     \toprule
                & \multicolumn{2}{c|}{Self Persona} & \multicolumn{2}{c}{Partner Persona} \\
     \hline
                & $\textbf{hits}@1$ & \textbf{MRR}  & $\textbf{hits}@1$ & \textbf{MRR}     \\
     \hline
      HRE-NA    & 47.4$^{\star}$  & 63.7$^{\star}$ & 42.2$^{\star}$ & 59.3$^{\star}$ \\
      HRE-CA    & 47.0$^{\star}$  & 63.7$^{\star}$ & 42.1$^{\star}$ & 59.3$^{\star}$ \\
      HRE-RA    & \textbf{58.1}$^{\star}$ & \textbf{71.8}$^{\star}$ & \textbf{42.8} & \textbf{60.0} \\
      HRE-CRA   & 43.3$^{\star}$  & 60.4           & 42.1$^{\star}$ & 59.1$^{\star}$ \\
     \hline
      IMN-NA    & 64.4$^{\star}$  & 76.3$^{\star}$ & 64.1           & 76.1  \\
      IMN-CA    & 64.6$^{\star}$  & 76.5$^{\star}$ & 63.9           & 76.1  \\
      IMN-RA    & \textbf{66.3}$^{\star}$  & \textbf{77.7}$^{\star}$ & \textbf{64.3}$^{\star}$ & \textbf{76.2}$^{\star}$  \\
      IMN-CRA   & 64.1            & 76.2           & 64.1           & 76.1  \\
     \hline
      BERT-NA   & 71.1            & 80.9           & 70.9           & 80.8  \\
      BERT-CA   & 71.2$^{\star}$  & 81.0           & 70.9           & 80.9  \\
      BERT-RA   & 82.6$^{\star}$  & 89.0$^{\star}$ & 71.1$^{\star}$ & 80.9  \\
      BERT-CRA  & \textbf{84.3}$^{\star}$ & \textbf{90.3}$^{\star}$ & \textbf{71.2}$^{\star}$ & \textbf{80.9}  \\
      \bottomrule
      \end{tabular}
      \label{tab3}
    \end{table}

    \begin{table*}[t]
     \caption{Performance of the proposed and previous methods on the \texttt{Persona-Chat} dataset under various persona configurations. 
     The meanings of ``Self Persona", ``Partner Persona", ``Original", and ``Revised" can be found in Section~\ref{sec2}. 
     The results of P$^2$ Bot~\cite{liu2020you} was reported on the validation set. 
     ``-" denotes that the results were not reported in their papers. 
     Numbers marked with $\star$ denote that the improvement over the best performing baseline is statistically significant (t-test with \emph{p}-value $<$ 0.05). 
     Numbers in bold denote the persona fusion strategy that achieves the best performance.} 
     \centering
     \begin{tabular}{l|c|c|c|c|c|c|c|c}
      \toprule
                                                         & \multicolumn{4}{c|}{Self Persona}                            & \multicolumn{4}{c}{Partner Persona} \\
      \hline
                                                         & \multicolumn{2}{c|}{Original} & \multicolumn{2}{c|}{Revised} & \multicolumn{2}{c|}{Original} & \multicolumn{2}{c}{Revised} \\
      \hline
                                                         & $\textbf{hits}@1$ & \textbf{MRR} & $\textbf{hits}@1$ & \textbf{MRR} & $\textbf{hits}@1$ & \textbf{MRR} & $\textbf{hits}@1$ & \textbf{MRR} \\
      \hline
       IR baseline ~\cite{zhang2018personalizing}        & 41.0   & -       & 20.7  & -      & 18.1  & -        & 18.1 & -           \\
       Starspace ~\cite{zhang2018personalizing}          & 48.1   & -       & 32.2  & -      & 24.5  & -        & 26.1 & -           \\
       Profile ~\cite{zhang2018personalizing}            & 47.3   & -       & 35.4  & -      & 28.3  & -        & 29.4 & -           \\
       KV Profile ~\cite{zhang2018personalizing}         & 51.1   & -       & 35.1  & -      & 29.1  & -        & 28.9 & -           \\
       FT-PC ~\cite{mazare2018training}                  & -      & -       & 60.7  & -      & -     & -        & -    & -           \\
       DGMN~\cite{zhao2019document}                      & 67.6   & -       & 58.8  & -      & -     & -        & -     & -     \\
       DIM~\cite{gu2019dually}                           & 78.8   & 86.7    & 70.7  & 81.2   & 64.0  & 76.1     & 63.9  & 76.0  \\
       TransferTransfo~\cite{wolf2019transfertransfo}    & 80.7   & -       & -     & -      & -     & -        & -     & -     \\
       P$^2$ Bot~\cite{liu2020you}                       & 81.9   & -       & 68.6  & -      & -     & -        & -     & -     \\
       FIRE~\cite{gu2020filtering}                       & 81.6   & -       & 74.8  & -      & -     & -        & -     & -     \\
      \hline
       BERT-RA                                           & 82.6$^{\star}$   & 89.0$^{\star}$    & 77.1$^{\star}$  & 85.4$^{\star}$   & 71.1$^{\star}$  & 80.9$^{\star}$     & 70.8$^{\star}$  & 80.8$^{\star}$ \\
       BERT-CRA                                          & \textbf{84.3}$^{\star}$  & \textbf{90.3}$^{\star}$   & \textbf{79.4}$^{\star}$  & \textbf{86.9}$^{\star}$   & \textbf{71.2}$^{\star}$  & \textbf{80.9}$^{\star}$    & \textbf{71.8}$^{\star}$ & \textbf{81.5}$^{\star}$ \\
       BERT-CRA - subtype                                & 83.6$^{\star}$   & 89.9$^{\star}$    & 78.4$^{\star}$  & 86.4$^{\star}$   & 70.8$^{\star}$   & 80.8$^{\star}$    & 70.9$^{\star}$  & 80.8$^{\star}$   \\
      \bottomrule
      \end{tabular}
      \label{tab2}
    \end{table*}

    For comparison, Table~\ref{tab1} presents the performance of HRE, IMN and BERT we reimplemented in this paper together with previous methods on this task without using any personas. 
    In our experiments, all reported results were the top one out of four runs under the same model configuration. 
    Thus, the reported baseline results were the best we attain.\footnote{Our reimplemented HRE, IMN and BERT were the same as those presented in \cite{serban2016building,gu2019interactive,devlin2019bert} on their datasets.}
    
    Table~\ref{tab3} presents the evaluation results of our proposed persona fusion strategies implemented into three models on the \texttt{Persona-Chat} dataset conditioned on the original persona.
    We can draw several conclusions from Table~\ref{tab3} as follows.
    First, self personas are more important than partner personas under all strategy and model configurations.
    It is reasonable that the self persona provides the fundamental descriptions of the speaker who is about to utter a response.
    Second, the partner persona was shown to contribute to the performance.
    Particularly our study demonstrates it is beneficial in the cross-attention-based and pretraining-based models.
    The reason might be that the interactions between contexts and responses could help to determine the importance of different parts of a persona and then fuse them.
    Third, the RA persona fusion strategy performs best among four strategies in the sentence-encoding-based and cross-attention-based models and the CRA fusion strategy performs best in the pretraining-based model.
    We consider this is because responses are closely related to personas while pretraining algorithms seem to help capture more semantics given additional contexts.

    Table~\ref{tab2} presents the evaluation results of our proposed and previous methods on the \texttt{Persona-Chat} dataset under various persona configurations.
    Our BERT-based model implemented with the context-response-aware persona fusion strategy achieves a new state-of-the-art performance.
    We can see that incorporating the generic distributional semantics and external knowledge learned from pretraining rendered improvements on both $\textbf{hits}@1$ and \textbf{MRR} conditioned on various personas.
    Compared with the FT-PC model \cite{mazare2018training}, BERT-CRA and BERT-RA outperformed it by margins of 18.7\% and 16.4\% respectively in terms of $\textbf{hits}@1$ conditioned on revised self personas.
    The results show that the generic distributional semantics and other knowledge learning from pretraining is beneficial for building personalized dialogue agents.
    Compared with TransferTransfo \cite{wolf2019transfertransfo} and P$^2$ Bot~\cite{liu2020you} which were also equipped with the generic distributional semantics, BERT-CRA and BERT-RA still outperformed them, which shows the effectiveness of our proposed persona fusion strategies.
    Lastly, BERT-CRA and BERT-RA outperformed all previous pretraining-based and non-pretraining-based methods by margins larger than 2.7\% and 1.0\% respectively $\textbf{hits}@1$ conditioned on original self personas, and margins larger than 4.6\% and 2.3\% respectively in terms of $\textbf{hits}@1$ conditioned on revised self personas.
    The results show that our proposed models achieved superiority on original personas and greater advantages on revised personas, achieving a new state-of-the-art performance of response selection on the \texttt{Persona-Chat} dataset.

  \subsection{Analysis}
  
    \paragraph{Subtype Embeddings}
    In order to demonstrate the importance of subtype embeddings used in BERT-CRA, an ablation test was further performed and the results were shown in the last row in Table~\ref{tab2}.
    We can see that the subtype embeddings contribute to the performance of BERT-CRA, which shows its effectiveness to distinguish personas, contexts and responses from each other.
    
    \begin{table}[t]
      \caption{The efficiency (cases/second) of four persona fusion strategies implemented into three models by recordding the inference time over the whole validation set on the \texttt{Persona-Chat} dataset under the original persona configuration.}
      \centering
      \begin{tabular}{lc|lc|lc}
      \toprule
                 \multicolumn{6}{c}{Efficiency (cases/second)}   \\
      \midrule
       HRE-NA    & 4660.1  &  IMN-NA   & 1661.4  &  BERT-NA  & 53.29 \\
       HRE-CA    & 4596.9  &  IMN-CA   & 1666.3  &  BERT-CA  & 53.29 \\
       HRE-RA    & 4626.9  &  IMN-RA   & 1674.6  &  BERT-RA  & 53.29 \\
       HRE-CRA   & 4643.5  &  IMN-CRA  & 1688.0  &  BERT-CRA & 92.67 \\
      \bottomrule
      \end{tabular}
      \label{tab8}
    \end{table}
    
    \paragraph{Retrieval Time}
    The retrieval time is very critical for retrieval-based chatbots. 
    Thus, we tested the time complexity by recording the inference time over the whole validation set using a GeForce RTX 2080 Ti GPU. 
    The results were reported by averaging two runs as shown in Table~\ref{tab8}. 
    Although BERT-based models take more time, it is acceptable compared to the performance they achieved. 
    Meanwhile, some recent studies have explored the methods of accelerating inference of BERT for the deployment in real-time applications. 
    This is a bit out of the scope of this paper, and will be explored in the future.

  \subsection{Discussion on Response Generation}
    Although fusing personas for dialogue generation is not the focus of this paper, we conducted a preliminary experiment to show that self or partner personas also contribute differently to response generation. 
    It is notable that only context is available while response candidate is not during inference in response generation.
    Thus in this section, we explored the impact of self and partner personas on only the context-aware persona fusion strategy.
    This strategy was implemented into pretraining-based models.
    In our experiment, we adopt a lightweight model MiniLM \cite{wang2020minilm} for the consideration of time and space complexity.
    Due to space limitation, we omit the introduction of MiniLM and readers can refer to \citet{wang2020minilm}.

    \begin{figure}[t]
      \includegraphics[width=7.5cm]{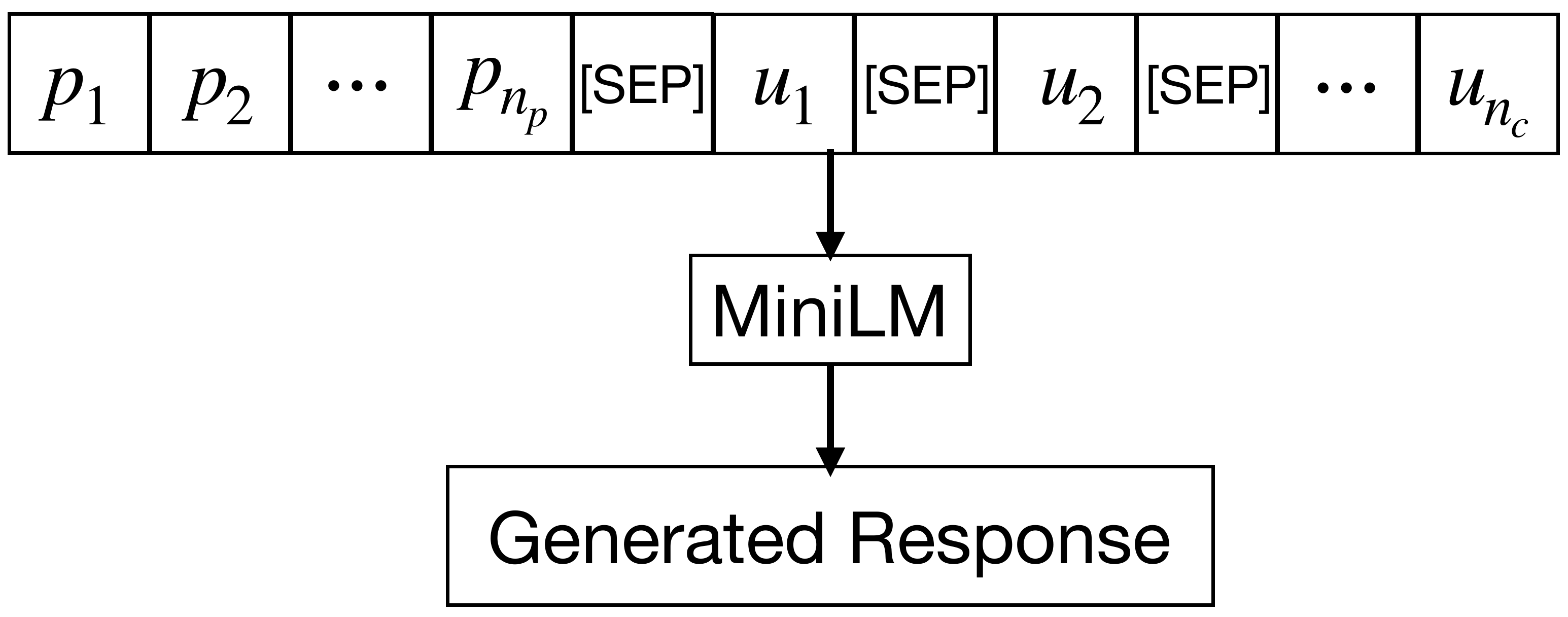}
      \caption{An overview of the model architecture for MiniLM with the context-aware persona fusion strategy.}
      \label{fig3}
    \end{figure}

    The overview of the model architecture for MiniLM with the context-aware persona fusion strategy is shown in Figure~\ref{fig3}.
    In this strategy, we follow the setting in previous studies \cite{zheng2019personalized,liu2020you} in which all the persona profiles are concatenated directly and the context utterances are concatenated and separated with the \texttt{[SEP]} token.
    Then, the persona and the context are concatenated with a \texttt{[SEP]} token.
    The concatenated persona-context combination are fed into the model as the input for the Seq2Seq generation.

    \begin{table}[t]
      \caption{Performance of response generation conditioned on the original persona. 
      Numbers marked with $\star$ denote that the gains or losses after adding persona conditions are statistically significant (t-test with \emph{p}-value $<$ 0.05).
      Numbers in bold denote the persona fusion strategy that achieves the best performance.}
      \centering
      \begin{tabular}{c|c|c|c|c}
      \toprule
        \multirow{2}{*}{Model-Persona}     & Relevance     & \multicolumn{2}{c|}{Diversity}    & \multirow{2}{*}{Length} \\
      \cline{2-4}
                                           & \textsc{Bleu} & \textsc{Dist-1} & \textsc{Dist-2} & \\
      \hline
        MiniLM                             &    3.54       &     94.47       &     99.49       & 8.73 \\
        MiniLM-CA-Self                     & \textbf{4.50}$^{\star}$ & \textbf{94.94}$^{\star}$  & \textbf{99.59}  & 8.59 \\
        MiniLM-CA-Partner                  &    3.65$^{\star}$       &     93.80$^{\star}$       &     99.48       & 9.05$^{\star}$ \\
      \bottomrule
      \end{tabular}
      \label{tab5}
    \end{table}
    
    We take two sets of widely used metrics to evaluate the relevance and the diversity of the generated responses. 
    For relevance, we use BLEU \cite{papineni2002bleu} which is a weighted summation of BLEU 1-4 and the length. 
    For diversity, we calculate the uni-gram and bi-gram distinct ratios (DIST-1, DIST-2) at the instance level \cite{li2016diversity}.
    
    Table~\ref{tab5} presents the performance of MiniLM with context-aware (CA) persona fusion strategy on the task of response generation using the  original version of persona.
    As we can see, the conclusions are consistent with those on response selection.
    First, compared to partner persona, self persona can help achieve better results on both the relevance and diversity metrics, which demonstrates that self persona can provide more fundamental information about the speaker who is about to utter a response.
    Second, although the partner persona was mostly thought to be not useful in previous studies, our results show that partner persona contribute to the performance on response generation.
    When partner persona is given, the BLEU score is improved on the MiniLM-based models.
    
    \begin{table}[t]
      \caption{An example from the generated responses that demonstrates the different contributions of the self and partner personas. Given the conversation context, \XSolidBrush denotes an inappropriate response and \CheckmarkBold denotes an appropriate one.}
      \centering
      \begin{tabular}{l|l}
      \toprule
       \textbf{Self Persona} & \textbf{Partner Persona} \\
      \midrule
       I like to \textcolor{green}{go hunting}.  &  I live in \textcolor{red}{Ohio}. \\
       I am a handyman.                          &  I like to \textcolor{blue}{go hiking}. \\
       I am allergic to shellfish.               &  I am a single mom of two boys. \\
       I restore classic cars.                   &  I work as an accountant. \\
      \midrule
       \multicolumn{2}{l}{\textbf{Context:}} \\
       \multicolumn{2}{l}{how are you tonight , i just got back from hiking .} \\
       \multicolumn{2}{l}{\textbf{Response with self persona:}} \\
       \multicolumn{2}{l}{i am good . i just got back from \textcolor{green}{hunting} .} \\
       \multicolumn{2}{l}{\textbf{Response with partner persona:}} \\
       \multicolumn{2}{l}{that sounds fun . i just got back from \textcolor{blue}{a hike}. \XSolidBrush} \\
       \multicolumn{2}{l}{Hiking in \textcolor{red}{Ohio} must be very interesting.  \CheckmarkBold} \\
      \bottomrule
      \end{tabular}
      \label{tab7}
    \end{table}
    
    We assume that self and partner personas contribute differently in response generation, while an example shown in Table~\ref{tab7} can verify our assumption to some extent. 
    As we can see, if we consider the partner persona as the self persona equally, the generated response will confuse the information from self or partner. 
    Designing effective personas fusion strategies for self and partner on the task of response generation is a large scope and we will leave it to our future work.

\section{Conclusions}
  In this paper, we propose four persona fusion strategies to explore the impact of self and partner personas on personalized response selection in retrieval-based chatbots.
  These strategies are implemented into three representative models for evaluation and comparison.
  Empirical studies on the \texttt{Persona-Chat} dataset show that the partner persona neglected in previous studies can still improve the performance under certain conditions.
  Besides, our proposed models improve the accuracy of response selection, outperforming previous methods by large margins and achieving a new state-of-the-art performance of response selection on the \texttt{Persona-Chat} dataset. 
  In the future, we will work on exploring the impact of self and partner personas for dialogue response generation to further verify the usefulness of partner in dialogue.

\section*{Acknowledgements}
  We thank anonymous reviewers for their valuable comments. 


\bibliographystyle{ACM-Reference-Format}
\bibliography{sample-base}


\end{document}